\def\year{2021}\relax
\documentclass[letterpaper]{article} 
\usepackage{aaai21}  
\usepackage{times}  
\usepackage{helvet} 
\usepackage{courier}  
\usepackage[hyphens]{url}  
\usepackage{graphicx} 
\usepackage{natbib}  
\usepackage{caption} 
\usepackage{multirow}

\usepackage{algorithm}
\usepackage{algorithmicx}
\usepackage[noend]{algpseudocode}

\usepackage{amsmath}
\usepackage{amssymb}
\usepackage{amsfonts}
\newcommand{\E}{\mathbb{E}}
\newcommand{\Ls}{\mathcal{L}}

\newcommand{\Cov}{\mathrm{Cov}}

\DeclareMathOperator*{\argmax}{arg\,max}
\DeclareMathOperator*{\argmin}{arg\,min}

\algdef{SE}[DOWHILE]{Do}{doWhile}{\algorithmicdo}[1]{\algorithmicwhile\ #1}

\usepackage{pifont}

\newcommand{\xmark}{\text{\ding{55}}}

\usepackage{hyperref}
\usepackage{xcolor}

\title{Curriculum Labeling: Revisiting Pseudo-Labeling for Semi-Supervised Learning}
\author {
    Paola Cascante-Bonilla, 
    Fuwen Tan, 
    Yanjun Qi, 
    Vicente Ordonez \\
}
\affiliations {
    University of Virginia \\
    $\{$pc9za, fuwen.tan, yanjun, vicente$\}$@virginia.edu
}

\begin{document}

\maketitle

\begin{abstract}
In this paper we revisit the idea of pseudo-labeling in the context of semi-supervised learning where a learning algorithm has access to a small set of labeled samples and a large set of unlabeled samples. 
Pseudo-labeling works by applying pseudo-labels to samples in the unlabeled set by using a model trained on the combination of the labeled samples and any previously pseudo-labeled samples, and iteratively repeating this process in a self-training cycle. 
Current methods seem to have abandoned this approach in favor of consistency regularization methods that train models under a combination of different styles of self-supervised losses on the unlabeled samples and standard supervised losses on the labeled samples. 
We empirically demonstrate that pseudo-labeling can in fact be competitive with the state-of-the-art, while being more resilient to out-of-distribution samples in the unlabeled set. We identify two key factors that allow pseudo-labeling to achieve such remarkable results (1) applying curriculum learning principles and (2) avoiding concept drift by restarting model parameters before each self-training cycle. 
We obtain $94.91$\% accuracy on CIFAR-10 using only $4,000$ labeled samples, and $68.87$\% top-1 accuracy on Imagenet-ILSVRC using only
10\% of the labeled samples. The code is available at
\href{https://github.com/uvavision/Curriculum-Labeling}{\color{brown}{https://github.com/uvavision/Curriculum-Labeling}}.
\end{abstract}

\section{Introduction}
Access to annotated examples has been critical in achieving significant improvements in a wide range of computer vision tasks \cite{effectivenessData, revisitEffectivenessData}. 
However, annotated data is typically limited or expensive to obtain. Semi-supervised learning (SSL) methods promise to leverage unlabeled samples in addition to labeled examples to obtain gains in performance. 
Recent SSL methods for image classification have achieved remarkable results on standard datasets while only relying on a small subset of the labeled data and using the rest as unlabeled data~\cite{meanTeacher, VirtualAT, InterpolationCT, MixMatch}. 
These methods often optimize a combination of supervised and unsupervised objectives to leverage both labeled and unlabeled samples during training. 
Our work instead revisits the idea of pseudo-labeling where unlabeled samples are iteratively added into the training data by pseudo-labeling them with a weak model trained on a combination of labeled and pseudo-labeled samples. 
While pseudo labeling has been proposed before~\cite{PseudoLabel,Shi_2018_ECCV,labelPropagation,arazo2019pseudolabeling}, we revise it by using a self-paced curriculum that we refer to as curriculum labeling, and restarting the model parameters before every iteration in the pseudo-labeling process.
We empirically demonstrate through extensive experiments that
an implementation of pseudo-labeling trained under curriculum labeling, achieves comparable performance against many other recent methods.

When training a classifier, a common assumption is that the decision boundary should lie in low-density regions in order to improve generalization~\cite{LowDensitySep}; therefore unlabeled samples that lie either near or far from labeled samples should be more informative for decision boundary estimation. 
Pseudo-labeling generally works by iteratively propagating labels from labeled samples to unlabeled samples using the current model to re-label the data~\cite{ScudderST1965,FralickST1967,AgrawalaST1970,MITSSLChapelle}. 
Typically, a classifier is first trained with a small amount of labeled data and then used to estimate the labels for all the unlabeled data. High confident predictions are then added as training samples for the next step.
This procedure repeats for a specific number of steps or until the classifier cannot find more confident predictions on the unlabeled set. 
However, pseudo-labeling~\cite{PseudoLabel,labelPropagation,arazo2019pseudolabeling} has been largely surpassed by recent methods~\cite{UDA,MixMatch,Berthelot2020ReMixMatch}, and there is little recent work on using this approach for this task. 

\begin{figure*}[t]
\centering
  \includegraphics[width=.84\textwidth]{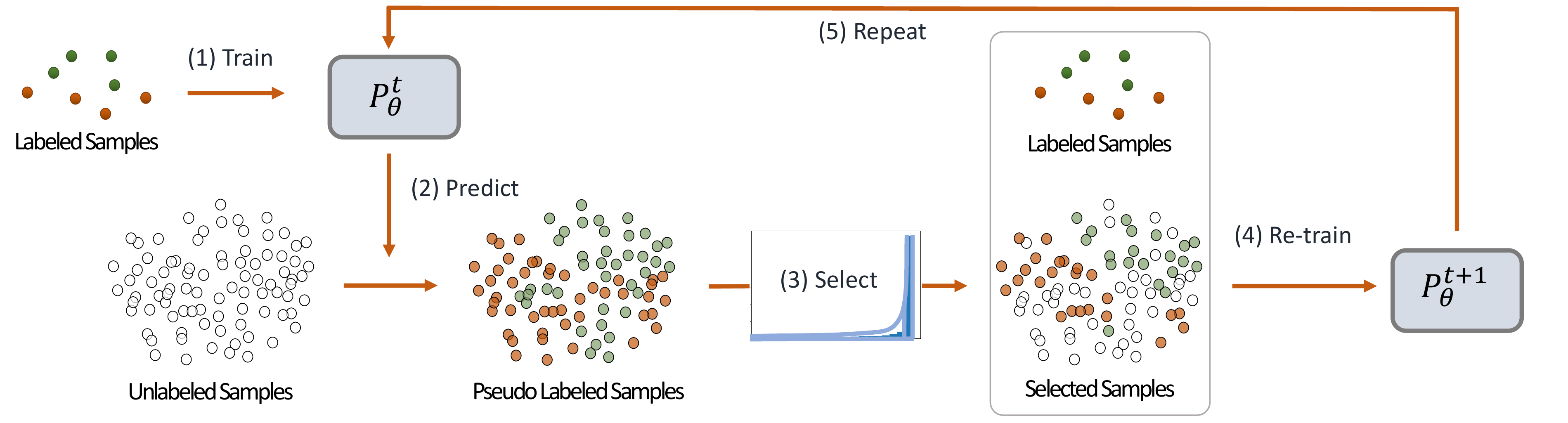}
\caption{\textbf{Curriculum Labeling (CL) algorithm.} 
The model is (1) trained on the labeled samples, then this model is used to (2) predict and assign pseudo-labels for the unlabeled samples. Then the distribution of the prediction scores is used to (3) select a subset of pseudo-labeled samples.  Then a new model is (4) re-trained with the labeled and pseudo-labeled samples. This process is (5) repeated by re-labeling unlabeled samples using this new model. The process stops when all samples in the dataset have been used during training.}
\label{fig:whole_model}
\vspace{-0.1in}
\end{figure*}

Our work borrows ideas from curriculum learning~\cite{CurriculumL} where a model first uses samples that are {\em easy} and progressively moves toward {\em hard} samples. 
Prior work has shown that training with a curriculum improves performance in several machine learning tasks~\cite{CurriculumL,curriculumDL}. 
The main challenge in designing a curriculum is how to control the pace -- going over the {\em easy} examples too fast may lead to more confusion than benefit while moving too slowly may lead to unproductive learning.
In particular, we show in our experiments, vanilla handpicked thresholding, as used in previous pseudo-labeling approaches~\cite{realisticEvaluation}, cannot guarantee success out of the box. 
Instead, we design a self-pacing strategy by analyzing the distribution of the prediction scores of the unlabeled samples and applying a criterion based on Extreme Value Theory (EVT)~\cite{Clifton_EVT2}. Figure~\ref{fig:whole_model} shows an overview of our model. We empirically demonstrate that curriculum labeling can achieve competitive results in comparison to the state-of-the-art, matching the recently proposed UDA~\cite{UDA} on Imagenet and surpassing UDA~\cite{UDA} and  ICT~\cite{InterpolationCT} on CIFAR-10.

A current issue in the evaluation of SSL methods was recently highlighted by~Oliver~et~al~\cite{realisticEvaluation}, which criticize the widespread practice of setting aside a small part of the training data as ``labeled'' and a large portion of the same training data as ``unlabeled''. 
This way of splitting the data leads to a scenario where all images in this type of ``unlabeled'' set have the implicit guarantee that they are drawn from the exact same class distribution as the ``labeled'' set. 
As a result, several SSL methods underperform under the more realistic scenario where unlabeled samples do not have this guarantee. We demonstrate that pseudo-labeling under a curriculum is surprisingly robust to out-of-distribution samples in the unlabeled set compared to other methods.

Our contributions can be summarized as follows:
\begin{itemize}
    \item We propose curriculum labeling (CL), which augments pseudo-labeling with a careful curriculum choice as pacing criteria based on Extreme Value Theory (EVT).

    \item We demonstrate that, with the proposed curriculum paradigm, the classic pseudo-labeling approach can deliver near state-of-the-art results on CIFAR-10, Imagenet ILSVRC top-1 accuracy, and SVHN -- compared to very recently proposed methods.

    \item Additionally, compared to previous approaches, our version of pseudo-labeling leads to more consistent results in realistic evaluation settings, where out-of-distribution samples are present. 
\end{itemize}

\section{Related Work}
\label{sec:related}
Our work is most closely related to the general use of pseudo-labeling in semi supervised learning~\cite{PseudoLabel, Shi_2018_ECCV, labelPropagation, arazo2019pseudolabeling}.
Although these methods have been surpassed by consistency regularization methods, our work suggests that this is not a fundamental flaw of self-training algorithms and demonstrate that careful selection of thresholds and pacing of the pseudo-labeling of the unlabeled samples can lead to significant gains. 
Here~\cite{PseudoLabel, Shi_2018_ECCV, arazo2019pseudolabeling} relied on a trained parametric model to pseudo-label unlabeled samples (e.g. by choosing the most confident class~\cite{PseudoLabel,Shi_2018_ECCV}, or using soft-labels~\cite{arazo2019pseudolabeling}), while~\cite{labelPropagation} proposed to propagate labels using a nearest-neighbors graph. Unlike these methods we adopt curriculum labeling which adds self-pacing to our training where we only add the most confident samples in the first few iterations and the threshold is chosen based on the distribution of scores on training samples.

Most closely related to our approach is the concurrent work of Arazo~et~al~\cite{arazo2019pseudolabeling} where pseudo-labeling is adopted in combination with Mixup~\cite{zhang2018mixup} to prevent what is referred in that work as confirmation bias. Confirmation bias is related in this context to the phenomenon of concept drift where the properties of a target variable change over time. This deserves special attention in pseudo-labeling since the target variables are affected by the same model feeding onto itself. 
In our version of pseudo-labeling we alleviate this confirmation bias by re-initializing the model before every self-training iteration, and as such, pseudo-labels from past epochs do not have an outsized effect on the final model. While our results are comparable on CIFAR-10, the work of Arazo~et~al~\cite{arazo2019pseudolabeling} further solidifies our finding that pseudo-labeling has a lot more to offer than previously found.

Recent methods for semi-supervised learning rather use a consistency regularization approach. In these works~\cite{Sajjadi2016RegularizationWS, TemporalEF, meanTeacher, Sohn2020FixMatchSS}, a network is trained to make consistent predictions in response to perturbation of unlabeled samples, by combining the standard cross-entropy loss with a consistency loss. 
Consistency losses generally encourage that in the absence of labels for a sample, its predictions should be consistent with the predictions for a perturbed version of the same sample. 
Various perturbation operations have been explored, including basic image processing operators~\cite{UDA}, dedicated operators such as MixUp~\cite{zhang2018mixup,Berthelot2020ReMixMatch}, learned transformations~\cite{autoAugment}, and adversarial perturbations~\cite{VirtualAT}. While we optionally leverage data augmentation, our use of it is the same as in the standard supervised setting, and our method does not enforce any consistency for pairs of samples.


\section{Method: Pseudo-Labeling under Curriculum}
\label{sec:method}
In semi-supervised learning (SSL), a dataset $D = \{x | x \in X\}$ is comprised of two disjoint subsets:
a labeled subset $D_{L}  = \{(x,y) | x \in X, y \in Y\}$ and an unlabeled subset $D_{UL} = \{x | x \in X\}$, where $x$ and $y$ denote the input and the corresponding label. Typically $|D_{L}| \ll |D_{UL}|$. 

Pseudo-labeling builds upon the general self-training framework~\cite{self_training}, where a model goes through multiple rounds of training by leveraging its own past predictions.
In the first round, the model is trained on the labeled set $D_{L}$.
In all subsequent rounds, the training data is the union of $D_{L}$ and a subset of $D_{UL}$ pseudo-labeled by the model in the previous round.
In round $t$, let the training samples be denoted as $(X_t, Y_t)$ and the current model as $P_\theta^t$, where $\theta$ are the model parameters, and $t \in \{1, \cdots, T\}$. 
After round $t$, an unlabeled subset $\bar{X_t}$ is added into $X_{t+1} := X_1 \cup \bar{X_t}$, and the new target set is defined as $Y_{t+1} := Y_1 \cup \bar{Y_t}$. Here $\bar{Y_t}$ represents the pseudo labels of $\bar{X_t}$, predicted by model $P_\theta^t$.
In this sense, the labels are ``propagated'' to the unlabeled data points via $P_\theta^t$.

The criterion used to decide which subset of samples in $D_{UL}$ to incorporate into the training in each round is key to our method. 
Different uncertainty metrics have been explored in the previous literature, including choosing the samples with the highest-confidence~\cite{Zhu06semi-supervisedlearning}, or retrieving the nearest samples in feature space~\cite{Shi_2018_ECCV, labelPropagation, SNTG}. 
We draw insights from Extreme Value Theory (EVT) which is often used to simulate extreme events in the tails of one-dimensional probability distributions by assessing the probability of events that are more extreme~\cite{adaptiveThresholdEVT,EVMachine,AlBehadili_EVT1,Clifton2008AutomatedND,paretoDist}. In our problem, we observed that the distribution of the maximum probability predictions for pseudo-labeled data follows this type of Pareto distribution. So instead of using fixed thresholds or tuning thresholds manually, we use percentile scores to decide which samples to add. Algorithm~\ref{alg:SS3T1L} shows the full pipeline of our model, where $\textit{Percentile}(X,T_r)$ returns the value of the $r$-th percentile. We use values of $r$ from 0\% to 100\% in increments of 20.

\begin{algorithm}[h!] 
\caption{Pseudo-Labeling under Curriculum Labeling}\label{alg:SS3T1L}
\begin{algorithmic}[1]
\footnotesize
\State \textbf{Require:} {$D_L$} \Comment{set of labeled samples}
\State \textbf{Require:} {$D_{UL}$} \Comment{set of unlabeled samples}
\State \textbf{Require:} {$\Delta := 20$} \Comment{stepping threshold percent }
\State {$P_\theta^t$} $\gets$ train classifier using {$D_L$} only
\State {$t := 1$}
\State {$T_r := 100 - \Delta$}
\Do 
\State {$T := \textit{Percentile}(P_\theta^t(D_{UL}), T_r)$}
\State {$X_{t} := D_{L}$}
\For{$x \in D_{UL}$}
\If{$P_\theta^t(x)$ $>$ $T$} 
\State { $X_{t} := X_{t} \cup {(x, \textit{pseudo-label}(P_\theta^t, x))}$}
\EndIf
\EndFor
\State {$P_\theta^t$} $\gets$ train classifier from  scratch using {$X_{t}$}
\State $t := t + 1$
\State $T_r := T_r - \Delta$
\doWhile{$|X_{t}| \neq |D_{L} + D_{UL}|$} 
\State \textbf{end}
\end{algorithmic}
\end{algorithm}

 Note that, in Algorithm~\ref{alg:SS3T1L} $\bar{X_t}$ is selected from the whole unlabeled set $D_{UL}$, enabling previous pseudo-annotated samples to enter or to leave the new training set. 
This is used to discourage concept drift or confirmation bias, as it can prevent erroneous labels predicted by an undertrained network during the early stages of training to be accumulated. 
To further alleviate the problem, we also reinitialize the model parameters $\theta$ with random initializations after each round, and empirically observe that, --as opposed to fine-tuning--, our reinitialization strategy leads to better performance.

Our termination criteria is that we keep iterating until all the samples in the pseudo-labeled set  comprise the entire training data samples which will take place when the percentile threshold is lowered to the minimum value ($T_r = 0$).

\section{Theoretical Analysis}
\label{sec:theory_main}

Our data consists of $N$ labeled examples $(X_i, Y_i)$ and $M$ examples $X_j$ which are unlabeled. Let $\mathcal{H}$ be a set of hypotheses $h_\theta$ where $h_\theta \in \mathcal{H}$, and each of them denotes a function mapping $X$ to $Y$.
Let $L_\theta(X_i)$ denote the loss for a given example $X_i$. 
To choose the best predictor with the lowest possible error, our formulation can be explained with a regularized Empirical Risk Minimization (ERM) framework.  Below, we define $\Ls(\theta)$ as the pseudo-labeling regularized empirical loss as:
\begin{small}
\begin{equation}
\begin{split}
\label{eq:ERM}
    & \mathcal{L}(\theta)=\hat\E[L_\theta]=\frac{1}{N}\sum_{i=1}^N L_\theta(X_i) + \frac{1}{M}\sum_{j=1}^M L'_\theta(X_j) \\ 
    & \hat\theta = \argmin_\theta \mathcal{L}(\theta) \\
    & L_\theta(X_i) = \texttt{CEE} (P_\theta^t(X_i), Y_i)\\   
    & L'_\theta(X_j) =\texttt{CEE} ( P_\theta^t(X_j), P_\theta^{t-1}(X_j))   
\end{split}
\end{equation}
\end{small}
Here \texttt{CEE} indicates cross entropy. Following \cite{curriculumDL} this can be rewritten as follows:
\begin{small}
\begin{equation}
\begin{split}
&    \hat\theta = \argmin_\theta \mathcal{L}(\theta) = \argmax_\theta \exp(-\mathcal{L}(\theta)) 
\end{split}
\end{equation}
\end{small}
Now to simplify the notions,  we reformulate the objective as to maximize a so-called \emph{Utility} objective, where $\delta$ is an indicator function: 
\begin{small}
\begin{equation}
\begin{split}
\label{eq:MU}
& U_\theta(X)=  e^{- \delta(X \in L) L_\theta(X) - \delta(X \in UL) L'_\theta(X) } \\
&    \mathcal{U}(\theta)=\hat\E[U_\theta]=\frac{1}{N}\sum_{i=1}^N U_\theta(X_i) + \frac{1}{M}\sum_{i=1}^M U'_\theta(X_j) \\ 
\end{split}
\end{equation}
\end{small}

The \emph{pacing function} in our pseudo curriculum labeling effectively provides a Bayesian prior for sampling unlabeled samples. This can be formalized as follows:

\begin{small}
\begin{align}
\label{eq:bayesian-utility}
&    \mathcal{U}_p(\theta)=\hat\E_p[U_\theta]=\frac{1}{N}\sum_{i=1}^N U_\theta(X_i) + \sum_{j=1}^M U'_\theta(X_j)p(X_j) \notag\\
    &=\frac{1}{N}\sum_{i=1}^N U_\theta(X_i) + \sum_{j=1}^M e^{-L'_\theta(X_j)}p_j \notag \\
&    \hat\theta = \argmax_\theta \mathcal{U}_p(\theta)
\end{align}
\end{small}

Here $p_j=p(X_j)$ denotes the induced prior probability we use to decide how to sample the unlabeled samples and is determined by the \emph{pacing function} of curriculum labeling algorithm, thus, $p(X_j)$ works like an indicator function where $1/M$ are the unlabeled samples whose scores are in the top $\mu$ percentile, and 0 otherwise.

We can then rewrite part of Eq~(\ref{eq:bayesian-utility}) as follows
\begin{small}
\begin{equation}
\label{eq:cov}
\begin{split}
    \mathcal{U}_p(\theta)&= \frac{1}{N}\sum_{i=1}^N U_\theta(X_i) + \sum_{j=1}^M (U'_\theta(X_j)-\hat\E[U'_\theta])(p_j-\hat\E[p]) \\
    & \hspace{0.45cm} + M * \hspace{1pt}\hat\E[U'_\theta]\hat\E[p ]\\ 
    & = \frac{1}{N}\sum_{i=1}^N U_\theta(X_i) +  \hat\Cov[U'_\theta,p] + M *  \hspace{1pt}\hat\E[U'_\theta]\hat\E[p] \\
    & =\mathcal{U}(\theta) + \hat\Cov[U'_\theta,p] 
\end{split}
\end{equation}
\end{small}

Eq~(\ref{eq:cov}) tells that when we sample the unlabeled data points according to a probability that positively correlates with the confidence of the predictions of $X_j$ (negative correlating with the loss $L'(X_j)$), we are improving the utility of the estimated $\theta$. Under such a pacing curriculum, we are minimizing the overall loss $\mathcal{L}$. The smaller the $L'(X_j)$ (the larger the utility of an unlabeled point $X_j$), the more likely we sample $X_j$. This theoretically proves that we want to sample more those unlabeled points that are predicted with more confidence. 

\section{Experiments}
\label{sec:experiments}

We first discuss our experiment settings in detail, 
then we compare against previous methods using the standard semi-supervised learning practice of setting aside a portion of the training data as labeled, and the rest as unlabeled, 
then we test in the more realistic scenario where unlabeled training samples do not follow the same distribution as labeled training samples, 
finally we conduct extensive evaluation justifying why our version of pseudo-labeling and specifically curriculum labeling enables superior results compared to prior efforts on pseudo-labeling. 

\subsection{Experimental Settings}
\label{sec:settings}
\textbf{Datasets:}
We evaluate the proposed approach on three image classification datasets: CIFAR-10~\cite{cifar10}, Street View House Numbers (SVHN)~\cite{svhn}, and ImageNet ILSVRC \cite{russakovsky2015imagenet,imagenet_cvpr09}.
With CIFAR-10 we use 4,000 labeled samples and 46,000 unlabeled samples for training and validation, and evaluate on 10,000 test samples. 
We also report results when the training set is restricted to 500, 1,000, and 2,000 labeled samples.
With SVHN we use 1,000 labeled samples and 71,257 unlabeled samples for training, 1,000 samples for validation, which is significantly lower than the conventional 7,325 samples generally used, and evaluate on 26,032 test samples. 
With ImageNet we use $\sim$10\% of the dataset as labeled samples (102,000 for training and 26,000 for validation), 1,253,167 unlabeled samples and 50,000 test samples.

\textbf{Model Details:}
We use CNN-13 \cite{CNN13_Base} and WideResNet-28 \cite{wideResidual} (depth 28, width 2) for CIFAR-10 and SVHN, and ResNet-50 \cite{resnet} for ImageNet. 
The networks are optimized using Stochastic Gradient Descent with nesterov momentum.
We use weight decay regularization of 0.0005, momentum factor of 0.9, and an initial learning rate of 0.1 which is then updated by cosine annealing~\cite{cosineAnnealing}.
Note that we use the same hyper-parameter setting for all of our experiments, except the batch size when applying moderate and heavy data augmentation. 
We empirically observe that small batches (i.e. 64-100) work better for moderate data augmentation (random cropping, padding, whitening and horizontal flipping), while large batches (i.e. 512-1024) work better for heavy data augmentation. For CIFAR-10 and SVHN, we train the models for 750 epochs. Starting from the 500th epoch, we also apply stochastic weight averaging (SWA)~\cite{SWA} every 5 epochs. 
For ImageNet, we train the network for 220 epochs and apply SWA, starting from the 100th epoch.

\textbf{Data Augmentation:}
While data augmentation has became a common practice in supervised learning especially when the training data is scarce \cite{dataAug1,dataAugEfectiveness,surveyDataAugmentation,dataAug2}, previous work on SSL typically use basic augmentation operations such as cropping, padding, whitening and horizontal flipping \cite{LadderNets,TemporalEF,meanTeacher}. 
More recent work relies on heavy data augmentation policies that are learned automatically by Reinforcement Learning \cite{autoAugment} or density matching \cite{FastAutoaugment}. 
Other augmentation techniques generate perturbations that take the form of adversarial examples \cite{VirtualAT}, or by interpolations between a random pair of image samples \cite{zhang2018mixup}. We explore both moderate and heavy data augmentation techniques that do not require to learn or search any policy, but instead apply transformations in an entirely random fashion. 
We show that using arbitrary transformations on the training set yields positive results. We refer to this technique as Random Augmentation (RA) in our experiments. However, we also report results without the use of data augmentation.

\begin{table*}[!th]
\scalebox{0.9}{
\begin{tabular}{l|l|c|c|c}
\hline
Approach & Method & ~~Year~~ & CIFAR-10 & SVHN \\ 
                        & & & $N_l$ = 4000 & $N_l$ = 1000 \\
\hline
 -- &Supervised              & -- & 20.26 $\pm$ 0.38 & 12.83 $\pm$ 0.47 \\
\hline
\multirow{2}{*}{\shortstack[l]{Pseudo\\Labeling}} & PL \cite{PseudoLabel}      & 2013 & ~17.78 $\pm$ 0.57~ &  ~7.62 $\pm$ 0.29~\\
&PL-CB~\cite{arazo2019pseudolabeling}               & 2019 & 6.28 $\pm$ 0.3 & - \\
\hline
\multirow{9}{*}{\shortstack[l]{Consistency\\ Regularization}}  & $\Pi$ Model \cite{TemporalEF}        & 2017 & 16.37 $\pm$ 0.63 &  7.19 $\pm$ 0.27\\
&Mean Teacher \cite{meanTeacher}      & 2017 & 15.87 $\pm$ 0.28 & 5.65 $\pm$ 0.47\\
&VAT \cite{VirtualAT}                 & 2018 & 13.86 $\pm$ 0.27 & 5.63 $\pm$ 0.20\\
&VAT + EntMin \cite{VirtualAT}        & 2018 & 13.13 $\pm$ 0.39 & 5.35 $\pm$ 0.19 \\
&LGA + VAT \cite{VAT_LGA}             & 2019 & 12.06 $\pm$ 0.19 & 6.58 $\pm$ 0.36 \\
&ICT \cite{InterpolationCT}           & 2019 & 7.66 $\pm$ 0.17 & 3.53 $\pm$ 0.07 \\
&MixMatch \cite{MixMatch}             & 2019 & 6.24 $\pm$ 0.06 & 3.27 $\pm$ 0.31 \\
&UDA \cite{UDA}                       & 2019 & 5.29 $\pm$ 0.25 & 2.46 $\pm$ 0.17 \\
&ReMixMatch \cite{Berthelot2020ReMixMatch} & 2020 & 5.14 $\pm$ 0.04 & 2.42 $\pm$ 0.09 \\
&FixMatch \cite{Sohn2020FixMatchSS} & 2020 & \textbf{4.26 $\pm$ 0.05} & \textbf{2.28 $\pm$ 0.11} \\
\hline
\multirow{4}{*}{\shortstack[l]{Pseudo\\Labeling}}&CL              & 2020 & 8.92 $\pm$ 0.03  &  5.65 $\pm$ 0.11 \\
&CL+FA\cite{FastAutoaugment}  & 2020 & 5.51 $\pm$ 0.14  & 2.90 $\pm$ 0.19 \\
&CL+FA\cite{FastAutoaugment}+Mixup\cite{zhang2018mixup}  & 2020 & 5.09 $\pm$ 0.18  & 2.75 $\pm$ 0.15 \\
&CL+RA+Mixup\cite{zhang2018mixup}    & 2020 & 5.27 $\pm$ 0.16  & 2.80 $\pm$ 0.18 \\
\hline
\end{tabular}}
\centering\caption{Test error rate on CIFAR-10 and SVHN using WideResNet-28. We show that our CL method can achieve comparable results to the state-of-the-art. "Supervised" refers to using only 4,000/1,000 labeled samples from CIFAR-10/SVHN without relying on any unlabeled data.}
\label{tab:WRNBaselines}
\end{table*}

\begin{table*}[!tbh]
\scalebox{0.9}{
 \begin{minipage}{.82\textwidth}
 \centering
\begin{tabular}{l|l|c|c}
\hline
Approach & Method & CIFAR-10  & SVHN\\ 
                                & & $N_l$ = 4000 & $N_l$ = 1000 \\
\hline
\multirow{2}{*}{\shortstack[l]{Pseudo\\ Labeling}}
&TSSDL-MT~\cite{Shi_2018_ECCV}        &  9.30 $\pm$ 0.55 & 3.35 $\pm$ 0.27\\
&LP-MT~\cite{labelPropagation}        & 10.61 $\pm$ 0.28 & -\\
\hline
\multirow{7}{*}{\shortstack[l]{Consistency\\ Regularization}}&Ladder net~\cite{LadderNets} & 12.36 $\pm$ 0.31 & -- \\
&MeanTeacher~\cite{meanTeacher} & 12.31 $\pm$ 0.24 & 3.95 $\pm$ 0.19 \\
&Temporal ensembling~\cite{TemporalEF} & 12.16 $\pm$ 0.24 & 4.42 $\pm$ 0.16 \\
&VAT~\cite{VirtualAT} & 11.36 $\pm$ 0.34 & 5.42\\
&VAT+EntMin~\cite{VirtualAT} & 10.55 $\pm$ 0.05 & 3.86 \\
&SNTG~\cite{SNTG} & 10.93  $\pm$ 0.14 & 3.86  $\pm$ 0.27 \\
&ICT~\cite{InterpolationCT} & 7.29 $\pm$ 0.02 & \textbf{2.89 $\pm$ 0.04} \\
\hline
\multirow{2}{*}{\shortstack[l]{Pseudo\\ Labeling}}&CL  & 9.81 $\pm$ 0.22  & 4.75 $\pm$ 0.28 \\
&CL+RA & \textbf{5.92 $\pm$ 0.07}  & 3.96 $\pm$ 0.10 \\
\hline
\end{tabular}
\end{minipage}
}
    \begin{minipage}{.16\textwidth}
    \caption{Test error rate on CIFAR-10 and SVHN using CNN-13. The value $N_l$ stands for the number of labeled examples in the training set.}
\label{tab:CNN13Baselines}
  \end{minipage}
   \end{table*}

\subsection{Comparisons with the State-of-the-Art}
\label{sec:state-of-the-art}
\textbf{CIFAR-10/SVHN:} In Table~\ref{tab:WRNBaselines} and~\ref{tab:CNN13Baselines}, we compare different versions of our method with state-of-the-art approaches using the WideResnet-28/CNN-13 architectures on CIFAR-10 and SVHN.
Our method surprisingly surpasses previous pseudo-labeling based methods~\cite{PseudoLabel, Shi_2018_ECCV, labelPropagation, arazo2019pseudolabeling} and consistency-regularization methods~\cite{UDA,MixMatch, Berthelot2020ReMixMatch} on CIFAR-10.
In SVHN we obtain competitive test error when compared with all previous methods that rely on moderate augmentation \cite{PseudoLabel,LadderNets, TemporalEF, meanTeacher, SNTG}, moderate-to-high data augmentation \cite{VirtualAT,VAT_LGA,InterpolationCT,MixMatch}, and heavy data augmentation \cite{UDA}.

\begin{figure}[tbh]
\centering
  \includegraphics[width=0.9\linewidth]{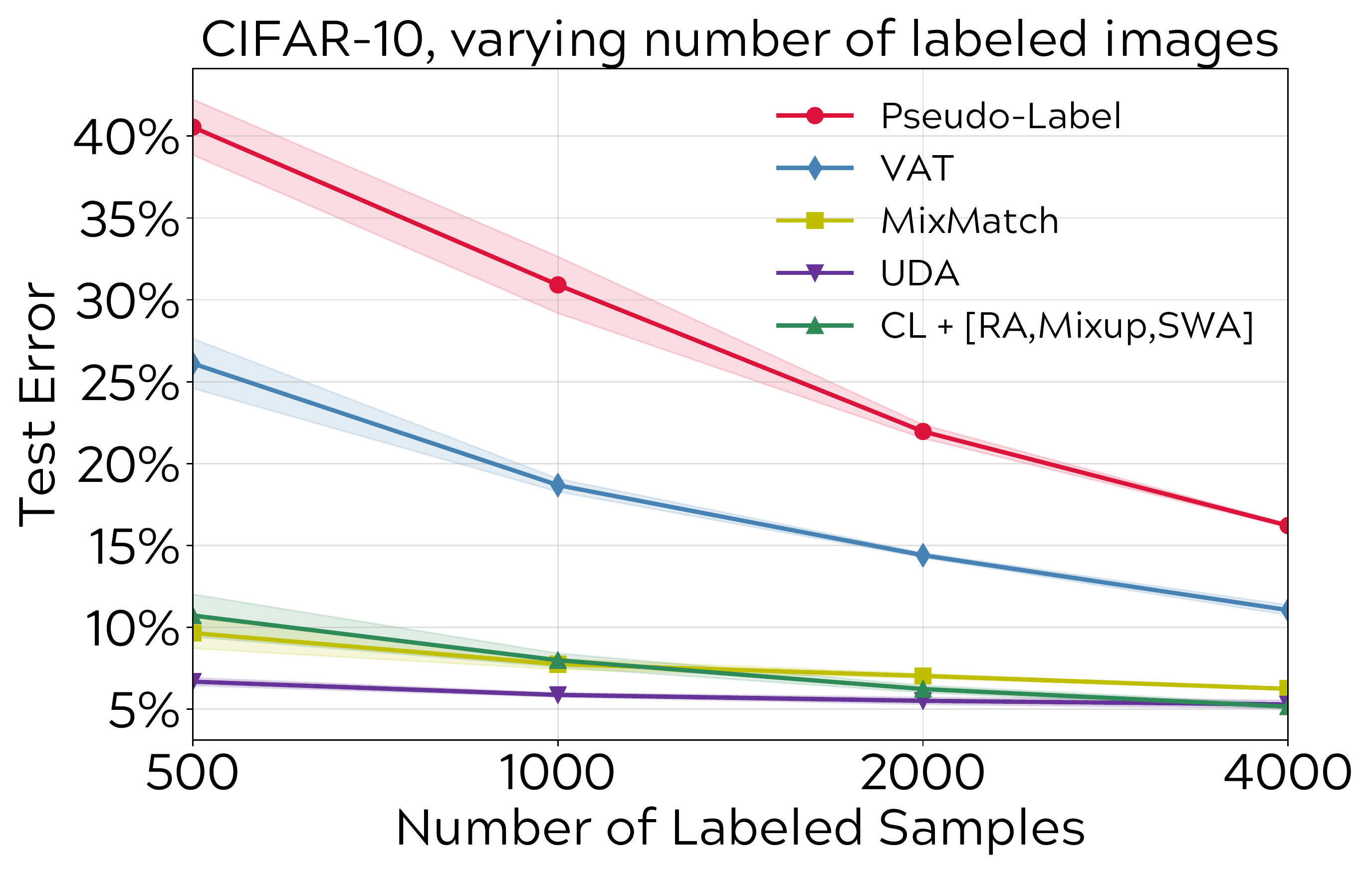}
\caption{Comparison of test error rate using WideResNet varying the size of the labeled samples on CIFAR-10. We use the standard validation set size of 5,000 to make our method comparable with previous work.}
\label{fig:num_labeled}
\end{figure}

A common practice to test SSL algorithms, is to vary the size of the labeled data using 50, 100 and 200 samples per class. 
In Figure~\ref{fig:num_labeled}, we also evaluate our method using this setting on WideResNet for CIFAR-10. We use the standard validation set size of 5,000 to make our method comparable with previous work. Decreasing the size of the available labeled samples recreates a more realistic scenario where there is less labeled data available.
We keep the same hyperparameters we use when training on 4,000 labeled samples, which shows that our model does not drastically degrade when dealing with smaller labeled sets. We show the lines for the mean and shaded regions for the standard deviation across five independent runs, and our results are closer to the current best method under this benchmark UDA~\cite{UDA}.

\textbf{ImageNet:} We further evaluate our method on the large-scale ImageNet dataset (ILSVRC). 
Following prior works~\cite{InterpolationCT,UDA,MixMatch}, we use 10\%/90\% of the training split as labeled/unlabeled data. Table \ref{tab:Imagenet_comp} shows that we achieve competitive results with the state-of-the-art with scores very close to the current top performing method, UDA~\cite{UDA} on both top-1 and top-5 accuracies.

\begin{table*}[tbh]
\scalebox{0.85}{
\begin{minipage}[t]{.75\textwidth}
\centering
\begin{tabular}{l|c|c|c}
\hline
Method & Approach & Top-1 & Top-5 \\
\hline
Supervised Baseline \cite{S4L} & -- & -- & 80.43 \\
Pseudo-Label \cite{PseudoLabel} & ~Pseudo Labeling~ & -- & 82.41 \\
VAT \cite{VirtualAT} & Consist. Reg. & -- & 82.78 \\
VAT + EntMin \cite{VirtualAT} & Consist. Reg. & -- & 83.39 \\
$S^{4}$ $L$-Rotation \cite{S4L} & ~Self-Supervision~ & -- & 83.82 \\
$S^{4}$ $L$-Exemplar \cite{S4L} & ~Self-Supervision~ & -- & 83.72 \\
\hline
UDA Supervised \cite{UDA} & -- & 55.09 & 77.26 \\ 
UDA Supervised (w. Aug) \cite{UDA}~  & -- & 58.84 & 80.56 \\
UDA (w. Aug) \cite{UDA} & Consist. Reg. & 68.78 & 88.80 \\
FixMatch \cite{Sohn2020FixMatchSS} & Consist. Reg. + PL & \textbf{71.46} & \textbf{89.13} \\
\hline
CL Supervised  (w. Aug) & -- & 55.75 & 79.67 \\
CL (w. Aug) & Pseudo Labeling & 68.87 & 88.56 \\
\hline
\end{tabular}
\end{minipage}}
\begin{minipage}{.21\textwidth}
\caption{Top-1 and top-5 accuracies on ImageNet with 10\% of the labeled set. 
Here UDA and CL are trained using ResNet-50. 
Previous methods \cite{PseudoLabel,VirtualAT,S4L} use ResNet-50v2 to report their results.  }
\label{tab:Imagenet_comp}
    \end{minipage}
\end{table*}

\subsection{Realistic Evaluation with Out-of-Distribution Unlabeled Samples} 
\label{sec:realistic-scenario}
\label{overlap}
\begin{figure}[!h]
\centering
  \includegraphics[width=0.9\linewidth]{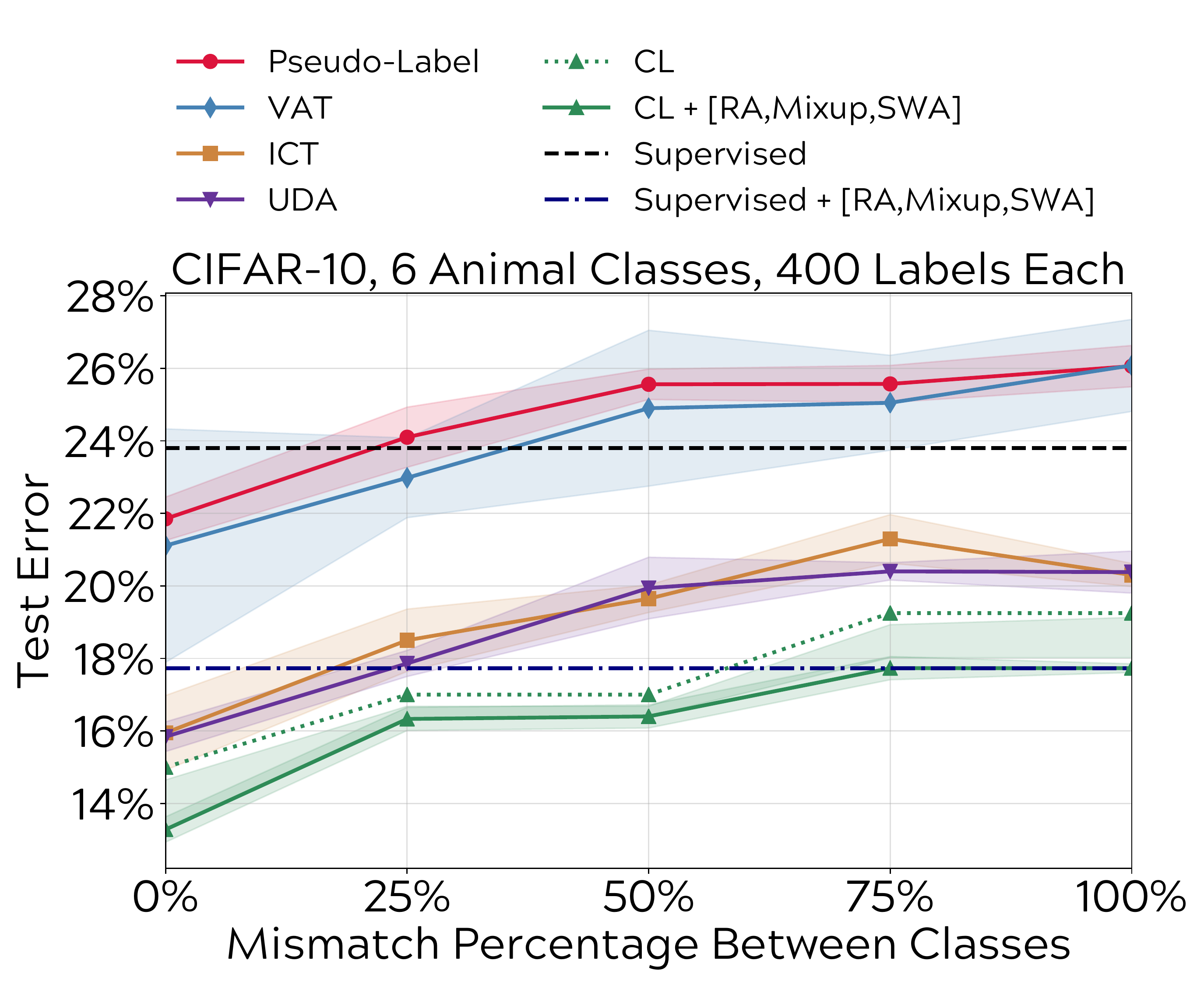}
\caption{Comparison of test error on CIFAR-10 (six animal classes) with varying overlap between classes. For example, in “50\%”, two of the four classes in the unlabeled data are not present in the labeled data. “Supervised” refers to using only the 2,400 labeled images.}
\label{fig:overlap}
\end{figure}

In a more realistic SSL setting~\cite{realisticEvaluation}, the unlabeled data may not share the same class set as the labeled data.
We test our method under a scenario where the labeled and unlabeled data come from the same underlying distribution, but the unlabeled data contains classes not present in the labeled data as proposed by \cite{realisticEvaluation}. 
We reproduce the experiment by synthetically varying the class overlap on CIFAR-10, choosing only the animal classes to perform the classification (bird, cat, deer, dog, frog, horse). 
In this setting, the unlabeled data comes from four classes. 
The idea is to vary how many of those classes are among the six animal classes to modulate the class distribution mismatch. 
We report the results of~\cite{PseudoLabel,VirtualAT} from~\cite{realisticEvaluation}. 
We also include the results of~\cite{InterpolationCT,UDA} obtained by running their released source code.  Figure~\ref{fig:overlap} shows that our method is robust to out-of-distribution classes, while the performance of previous methods drops significantly. 
We conjecture that our self-pacing curriculum is key to this scenario, where the adaptive thresholding scheme could help filter the out-of-distribution unlabeled samples during training.

\subsection{Ablation Studies}
\label{sec:ablation}
Here we justify the effectiveness of the two main design differences in our version of pseudo-labeling with respect to previous attempts. We first demonstrate that the choice of thresholds using percentiles as in curriculum labeling, has a large effect on the results compared to fixed thresholds
, then we show that training the model parameters from scratch in each round of self-training is more beneficial than fine-tuning over previous versions of the model. 

\begin{table*}[!h]
\parbox{.47\linewidth}
{
    \centering
    \scalebox{0.85}{
    \begin{tabular}{c|c|c|c}
    \hline
     Data Augmentation & Mixup & SWA & Top-1 Error \\
    \hline
     None & \xmark & \xmark & 38.47 \\
     None & \checkmark & \xmark & 34.03 \\
     None & \xmark & \checkmark & 37.35 \\
     None & \checkmark & \checkmark & 32.85 \\
     Moderate & \xmark & \xmark & 22.8 \\
     Moderate & \checkmark & \xmark & 16.11 \\
     Moderate & \xmark & \checkmark & 21.63 \\
     Moderate & \checkmark & \checkmark & 15.83 \\
     Heavy (RA) & \xmark & \xmark & 11.38 \\
     Heavy (RA) & \checkmark & \xmark & 8.88 \\
     Heavy (RA) & \xmark & \checkmark & 11.32 \\
     Heavy (RA) & \checkmark & \checkmark & 8.59 \\
    \hline
    \end{tabular}}
    \caption{Test errors when using pseudo-labeling without a curriculum (the threshold is set to 0.0). We use WideResnet-28 as the base network. Additionally, we report which data augmentation technique was applied.}
    \label{tab:no_thresholds}
}
\hfill
\parbox{.49\linewidth}
{
    \centering
    \scalebox{0.85}{
    \begin{tabular}{c|c|c}
    \hline
    Threshold & Moderate Aug & Heavy (RA) \\
    \hline
     0.1 & 21.12 & 7.87  \\
     0.2 & 21.12 & 8.57  \\
     0.3 & 20.59 & 7.90  \\
     0.4 & 22.11 & 8.15 \\
     0.5 & 19.98 & 7.46 \\
     0.6 & 19.51 & 6.88  \\
     0.7 & 19.35 & 6.65  \\
     0.8 & 18.08 & 6.29  \\
     0.9 & 17.11 &  6.21 \\
    \hline
     CL & 8.92 & 5.27 \\ 
    \hline
    \end{tabular}}
    \caption{Test errors when using pseudo-labeling with several fixed thresholds and different data augmentation techniques. We use WideResnet-28 as the base network. The Heavy Augmentation applies Random Augmentation, Mixup and SWA. Last row shows results of our approach under the proposed curriculum criteria.}
    \label{tab:thres_more_results}
}
\end{table*}

\subsubsection{Effectiveness of Curriculum Labeling}
\label{sec:ablation-thresholds}
\label{sec:efect_self_pacing}

\begin{table*}[!h]
\scalebox{0.85}{
\begin{tabular}{l||c|c||c|c||c|c}
\hline
 & 0.9 \textsuperscript{1} \# & WRN\textsuperscript{1} & 0.9995 \textsuperscript{2} \# & WRN\textsuperscript{2} & Self-pacing \# & WRN \\
\hline
Fully Supervised & - & 18.25 & - & 18.25 & - & 18.25 \\
1º Iteration & $\sim$35k & 15.25 & $\sim$13k & 17.18 & 8k &  15.41\\
2º Iteration & $\sim$41k & 14.53 & $\sim$14k & 15.2  & 16k &  11.55\\
3º Iteration & $\sim$43k & 13.91 & $\sim$14k & 14.64 & 24k &  10.83 \\
4º Iteration & $\sim$44k & 14.01 & $\sim$17k & 14.84 & 32k &  9.54 \\
5º Iteration & $\sim$44k & 12.92 & $\sim$15k & 15.29 & ~41k~ & \textbf{8.92} \\
\hline
\end{tabular}}
\centering\caption{Test errors when using two static thresholds (0.9\textsuperscript{1} and 0.9995\textsuperscript{2}) and our self-pacing training. We use WideResnet-28(WRN) as the base network. Fully Supervised refers to using only 4,000 labeled datapoints from CIFAR-10 without any unlabeled data.
The \# columns show the average numbers of images automatically selected for each iteration during training. 
We used ZCA preprocessing and moderate data augmentation on these experiments.} 
\label{tab:thres}
\vspace{-0.1in}
\end{table*}

\begin{table}[!th]
\resizebox{\columnwidth}{!}{
\scalebox{0.85}{
\begin{tabular}{l|c|c|c|c}
\hline
& Top  &  &  & Fine \\
& Confidence & \# & Reinitializing & Tuning \\
\hline
Fully Supervised & -  &  -   &  15.42 & 15.3 \\  
1º Iteration & $80\%$ & 8k    &  10.04 & 9.85 \\  
2º Iteration & $60\%$ & 16k  &  8.56  & 7.99  \\ 
3º Iteration & $40\%$ & 24k   &  7.03  & 7.20  \\ 
4º Iteration & $20\%$ & 32k   &  6.22  & 6.55  \\ 
5º Iteration & $0\%$  & 41k   &  5.41  & 6.42  \\ 
\hline
\end{tabular}}}
\centering\caption{Comparison of model reinitialization and finetuning, in each iteration of training. We observe that reinitializing the model performs consistently better in each iteration.}
\label{table:finetuning}
\vspace{-0.1in}
\end{table}

We first show results when applying vanilla pseudo-labeling with no curriculum, and without a specific threshold (i.e. 0.0). Table~\ref{tab:no_thresholds} shows the effect when applying different data augmentation techniques (Random Augmentation, Mixup and SWA). We show that only when heavy data augmentation is used, this approach is able to match our curriculum design without any data augmentation. This vanilla pseudo-labeling approach is similar to the one reported in previous literature~\cite{PseudoLabel}, and is not able to outperform recent work based on consistency regularization techniques.  
We also report experiments applying smaller thresholds in each iteration, and show our results in table~\ref{tab:thres_more_results}. Our curriculum design is able to yield a significant gain over the traditional pseudo-labeling approach that uses a fixed threshold even when heavy data augmentation is applied. This shows that our curriculum approach is able to alleviate the concept drift and confirmation bias.

Then we compare our self-pacing curriculum labeling with handpicked thresholding mimicking the experiments presented in Oliver~et~al~\cite{realisticEvaluation} and report more detailed results per iteration. 
As performed in this earlier work, we re-label only the most confident samples; in our experiments we fixed the thresholds to 0.9 and 0.9995. 
We test on CIFAR-10 with 4000 labeled samples, 
and use WideResnet-28 as the base network with moderate data augmentation.
As shown in Table~\ref{tab:thres}, using handpicked thresholds is sub-optimal. 
Especially when only the most confident samples are re-labeled as in~\cite{PseudoLabel}. Particularly, there is little to no improvement after the first iteration.
Our self-pacing model significantly outperforms fixed thresholding and continues making progress in different iterations.

\subsubsection{Effectiveness of Reinitializing vs Finetuning}
\label{sec:ablation-finetuning}
Due to the confirmation bias and concept drift, errors caused by high confident mis-labeling in early iterations may accumulate during multiple rounds of training.
Although our self-pacing sample selection encourages excluding incorrectly labeled samples in early iterations, this might still be an issue.
As such, we decided to reinitialize the model after each iteration, instead of finetuning the previous model. Reinitializing the model yields at least 1\% improvement and does not add a significant overhead to our self-paced approach, which is significantly faster than recent methods (additional experiments in the Appendix).
In Table~\ref{table:finetuning}, we compare the performance of model reinitializing and finetuning on CIFAR-10 on the 4,000 labeled training samples regime. We use WideResnet-28 as the base network and apply data augmentation during training.
It shows that reinitializing the model, as opposed to finetuning, indeed improves the accuracy significantly, demonstrating an alternative and perhaps simpler solution to alleviate the issue of confirmation bias~\cite{arazo2019pseudolabeling}.  

\section{Conclusion}
\label{sec:conclusion}

In this paper, we revisit pseudo-labeling in the context of semi-supervised learning and demonstrate comparable results with the current state-of-the-art that mostly relies on enforcing consistency on the predictions for unlabeled samples. As part of our version of pseudo-labeling, we propose curriculum labeling where unlabeled samples are chosen by using a threshold that accounts for the skew in the distribution of the prediction scores on the unlabeled samples. We additionally show that concept drift and confirmation bias can be mitigating by discarding the current model parameters before each epoch in the self-training loop of pseudo-labeling. We demonstrate our findings with strong empirical results on CIFAR-10, SVHN, and ImageNet ILSVRC. 

\vspace{0.06in}
\noindent {\bf Acknowledgments} Part of this work was supported by generous funding from SAP Research. We also thank anonymous reviewers for their feedback.

\bibliography{egbib}

\newpage
\appendix
\section{Appendix}
\subsection{Decision boundaries when applying Pseudo-Labeling and Curriculum Labeling}

\begin{figure*}[h]
\centering
\includegraphics[width=.95\textwidth]{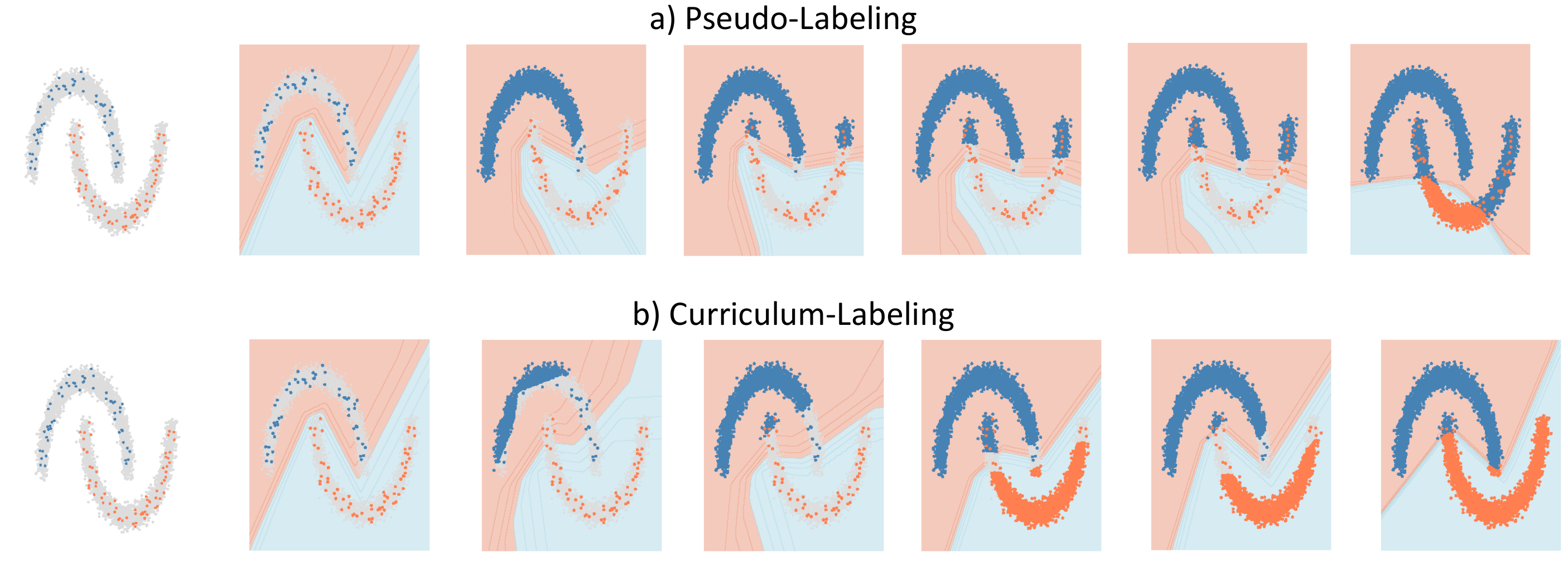}
\caption {Comparison of regular pseudo-labeling~(PL)~\cite{PseudoLabel} and pseudo-labeling with curriculum labeling~(CL) on the “two moons” synthetic dataset.
In blue are positive samples, in orange are negative samples, and in gray are unlabeled samples. Subsequent plots show how the learned decision boundaries evolve during training for the two approaches, showing that a self-paced approach leads to a more desirable state.
}
\label{fig:two_moon}
\vspace{0.15in}
\end{figure*}

We show in Figure~\ref{fig:two_moon} how curriculum labeling works on the synthetic two-moon dataset as it progressively labels the unlabeled samples. We observe that unlabeled samples get progressively labeled mostly with their correct labels as learning progresses.

\subsection{Theoretical Analysis Derivation}  \label{sec:theor_extra_eq5}

The derivation of Eq~(\ref{eq:cov}) in the Theoretical Analysis (section~\ref{sec:theory_main}) is as follows:
\begin{small}
\begin{equation}
\begin{split}
    \sum_{j=1}^M U'_\theta(X_j)p(X_j) = \sum_{j=1}^M U'_\theta(X_j)p(X_j) - M * \hat\E[U'_\theta]\hat\E[p] \\ 
    - M *  \hat\E[U'_\theta]\hat\E[p] + M *  \hat\E[U'_\theta]\hat\E[p] + M *  \hat\E[U'_\theta]\hat\E[p]\\
    = \sum_{j=1}^M U'_\theta(X_j)p(X_j) - \sum_{j=1}^M \hat\E[U'_\theta]p(X_j) \\
    - \sum_{j=1}^M U'_\theta(X_j)\hat\E[p] + \sum_{j=1}^M \hat\E[U'_\theta]\hat\E[p] + M *  \hat\E[U'_\theta]\hat\E[p]\\
    = \sum_{j=1}^M (U'_\theta(X_j)p(X_j) - \hat\E[U'_\theta]p(X_j) \\
    - U'_\theta(X_j)\hat\E[p] + \hat\E[U'_\theta]\hat\E[p]) + M *  \hat\E[U'_\theta]\hat\E[p]\\
    = \sum_{j=1}^M (U'_\theta(X_j)-\hat\E[U'_\theta])(p_j-\hat\E[p]) + M *  \hat\E[U'_\theta]\hat\E[p]
\end{split}
\end{equation}
\end{small}

\subsection{Hyperparameter Selection}

In table \ref{table:arch_res} we show the performance of our method when using data augmentation and we vary the batch size.
We keep the same hyperparameters across all experiments to ensure a fair comparison with the results obtained when training our method with a batch size of 512. We use CIFAR-10 with 4,000 labeled samples, using the rest of the training set as unlabeled samples, and report our results using the 10,000 test samples.

\begin{table*}[!h]
\begin{tabular}{l|c|c|c|c} \\ 
\hline
& Top & \# of Pseudo- & \multicolumn{2}{c}{Varying Batch Size} \\ 
& Confidence & Labeled Samples & Batch of 64 & Batch of 1024 \\ 
\hline
Fully Supervised & -  &  -  &  12.76  &  15.42 \\ 
1º Iteration & $80\%$ & 8k  &  10.19  &  10.04 \\ 
2º Iteration & $60\%$ & 16k &  9.32   &  8.56  \\ 
3º Iteration & $40\%$ & 24k &  8.49   &  7.03  \\ 
4º Iteration & $20\%$ & 32k &  7.34   &  6.22  \\ 
5º Iteration & $0\%$  & 41k &  7.17   &  5.41  \\ 
\end{tabular}
\centering\caption{Comparison of test error rate descent with our iterative method using two different batch sizes when using heavy data augmentation. Fully Supervised refers to using only 4,000 labeled datapoints from CIFAR-10 without any unlabeled data. Top Confidence, refers to the threshold applied in each Iteration, and \# of Pseudo-Labeled Samples is the average number of images automatically selected for each iteration when training.}
\label{table:arch_res}
\end{table*}

We also experiment using different percentiles, and found that using smaller percentiles lead to worse performance (about $\approx9.09$ test error), and higher percentiles (e.g. 5\%, 10\% and 15\%) lead to similar results as when using percentiles of 20\%. In table \ref{table:dif_thres} we show the performance of our method when using percentiles of 5\%. This experiment takes more than 48 hours to complete. We keep the same hyperparameters across all experiments to ensure a fair comparison with previous results. We use CIFAR-10 with 4,000 labeled samples, using the rest of the training set as unlabeled samples, and report our results using the 10,000 test samples.

\begin{table}[H]
\begin{tabular}{l|c|c|c}
\hline
  & Top & \# of Pseudo- & CIFAR-10 \\
  & Confid. & Labeled Sample & $N_t = 4000$ \\
\hline
Fully Sup. & -  &  -  &  ~~13.73~~ \\  
1º Iteration & $80\%$ & 8k  &  \underline{10.04}  \\  
2º Iteration & $75\%$ & 14k &  9.76   \\ 
3º Iteration & $70\%$ & 16k &  9.73 \\ 
4º Iteration & $65\%$ & 18k &  9.75 \\ 
5º Iteration & $60\%$ & 20k &  \underline{8.32} \\ 
6º Iteration & $55\%$ & 23k  &  8.54 \\  
7º Iteration & $50\%$ & 22k &  7.27 \\ 
8º Iteration & $45\%$ & 24k &  7.36 \\ 
9º Iteration & $40\%$ & 26k &  \underline{6.75} \\ 
10º Iteration & $35\%$ & 28k &  6.58 \\ 
11º Iteration & $30\%$ & 30k  &  5.89 \\  
12º Iteration & $25\%$ & 32k &  5.98 \\ 
13º Iteration & $20\%$ & 34k &  \underline{5.41} \\ 
14º Iteration & $15\%$ & 36k &  \textbf{5.27} \\ 
15º Iteration & $5\%$  & 38k &  5.45 \\ 
16º Iteration & $0\%$  & 41k &  \underline{5.31} \\ 
\end{tabular}
\centering\caption{Comparison of test error rate descent without using the Pareto Principle or 80/20 rule, but selecting a smaller stepping percentile. Fully Supervised refers to using only 4,000 labeled datapoints from CIFAR-10 without any unlabeled data. Top Confidence, refers to the threshold applied in each Iteration, and \# of Pseudo-Labeled Samples is the average number of images automatically selected for each iteration when training.}
\label{table:dif_thres}
\end{table}

In table \ref{table:epochs_swa} we show the performance of our method when varying the number of training epochs when training on WideResNet. We also show results when no stochastic weighting averaging is used. We use CIFAR-10 with 4,000 labeled samples, using the rest of the training set as unlabeled samples, and report our results using the 10,000 test samples.

\begin{table}[H]
\begin{tabular}{c|c|c|c }
\hline
Number of Epochs & Heavy (RA) & SWA & CIFAR-10 \\
& & & $N_l$ = 4000 \\
\hline
250  & \checkmark & \checkmark &  8.42 \\  
450  & \checkmark & \checkmark & 6.46 \\   
550  & \checkmark & \checkmark & 7.42 \\  
750  & \checkmark & $\times$ & 5.76  \\   
\end{tabular}
\centering\caption{Comparison of test error varying the number of training epochs and not applying SWA for CIFAR-10 with 4,000 labeled samples.}
\label{table:epochs_swa}
\end{table}

\subsection{Scalability and Computational Cost}  \label{sec:time_cost}

\begin{figure}[H]
\centering
  \includegraphics[width=.48\textwidth]{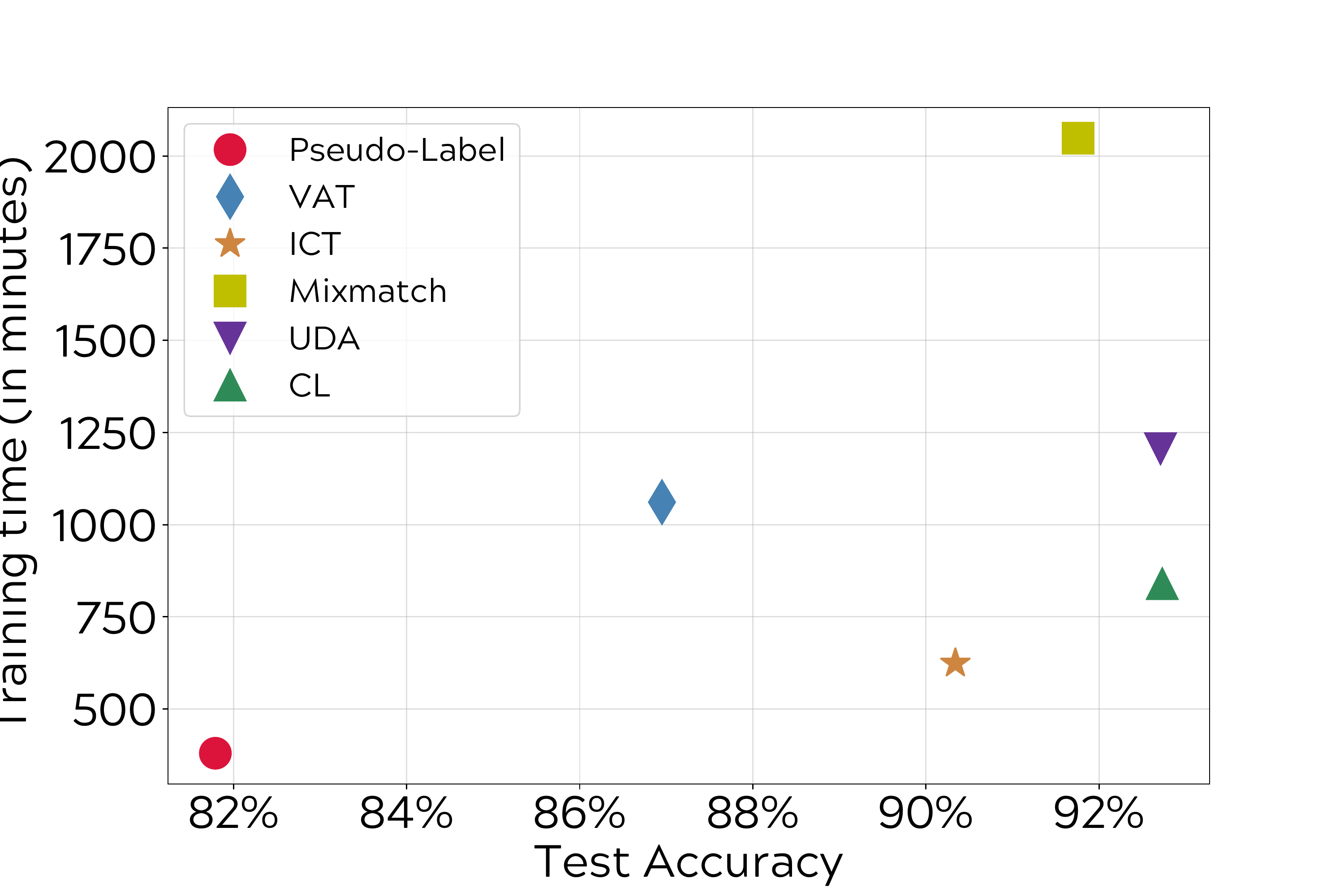}
  \vspace{-0.1in}
\caption{Comparison of time consumption and test accuracy on CIFAR-10 using 4,000 labeled samples. All the experiments were performed using Wide ResNet on an isolated virtual machine. The optimum condition between training time and accuracy is shown in the bottom right of the figure. }
\vspace{-0.15in}
\label{fig:timeCost}
\end{figure}

In figure \ref{fig:timeCost} we compare the time consumption versus the test accuracy on Wide ResNet for CIFAR-10 using 4,000 labeled samples. To report this results we perform our experiments on an isolated virtual machine on Google Cloud with 16 vCPUs and 1 Nvidia Tesla K80 GPU, and run each experiment separately. When considering both test accuracy and time cost, our approach shows strong advantages over other baselines. 

\subsection{Pseudo-Labeled Samples}
We examine some examples of pseudo-labeled images for CIFAR-10 and ImageNet, in order to provide a better intuition as to what are the typical types of examples that are being added during training. We include both some correctly and incorrectly pseudo-labeled examples, along with their prediction scores.

\begin{figure*}[th]
\centering
  \includegraphics[width=1\textwidth]{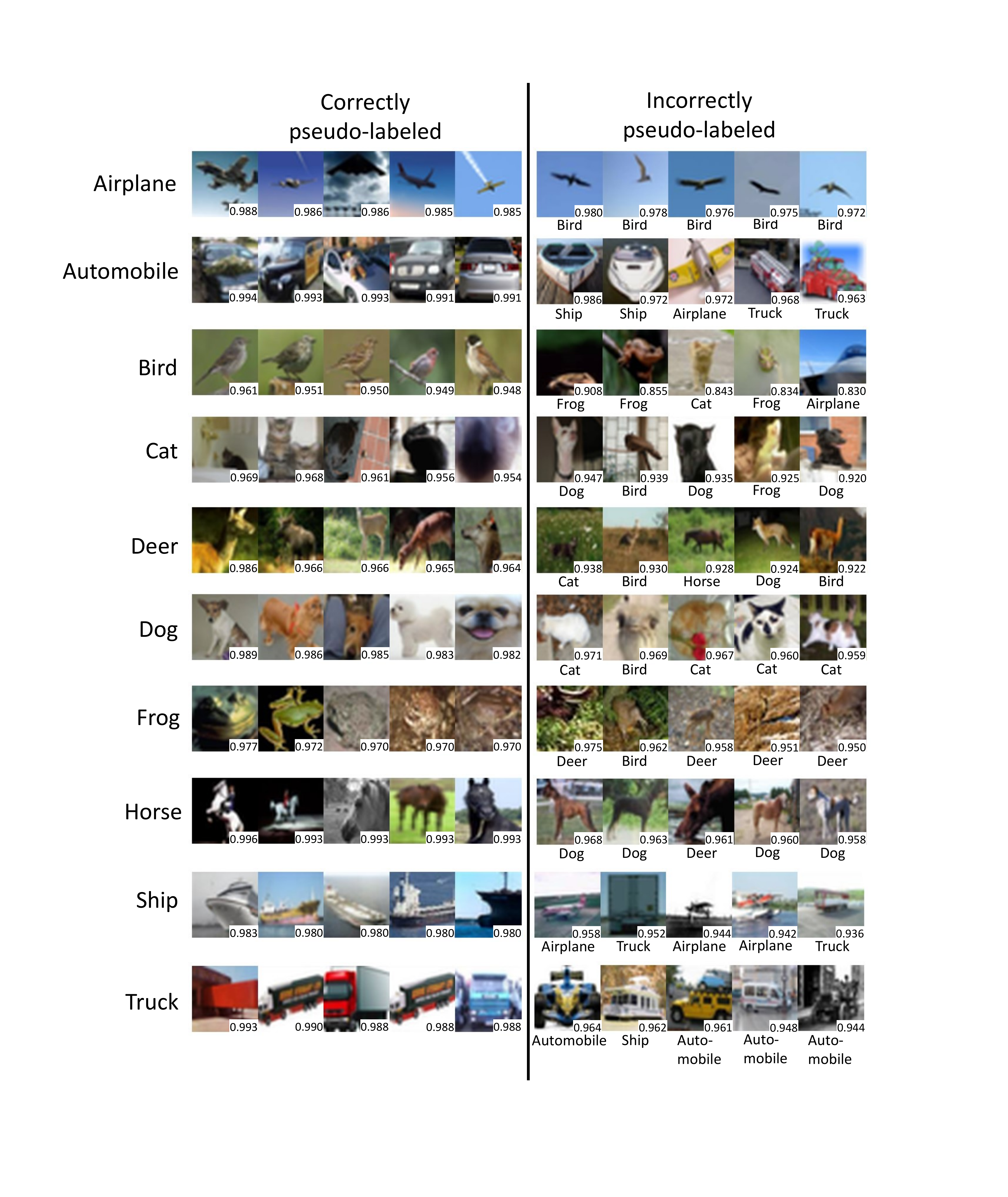}
  \vspace{-0.8in}
\caption{CIFAR-10 pseudo-labeled samples annotated by our method. We show correct and incorrect samples for the 10 categories with their corresponding scores after the last iteration.}
\label{fig:cifarSamples}
\end{figure*}

\begin{figure*}[th]
\centering  \includegraphics[width=1\textwidth]{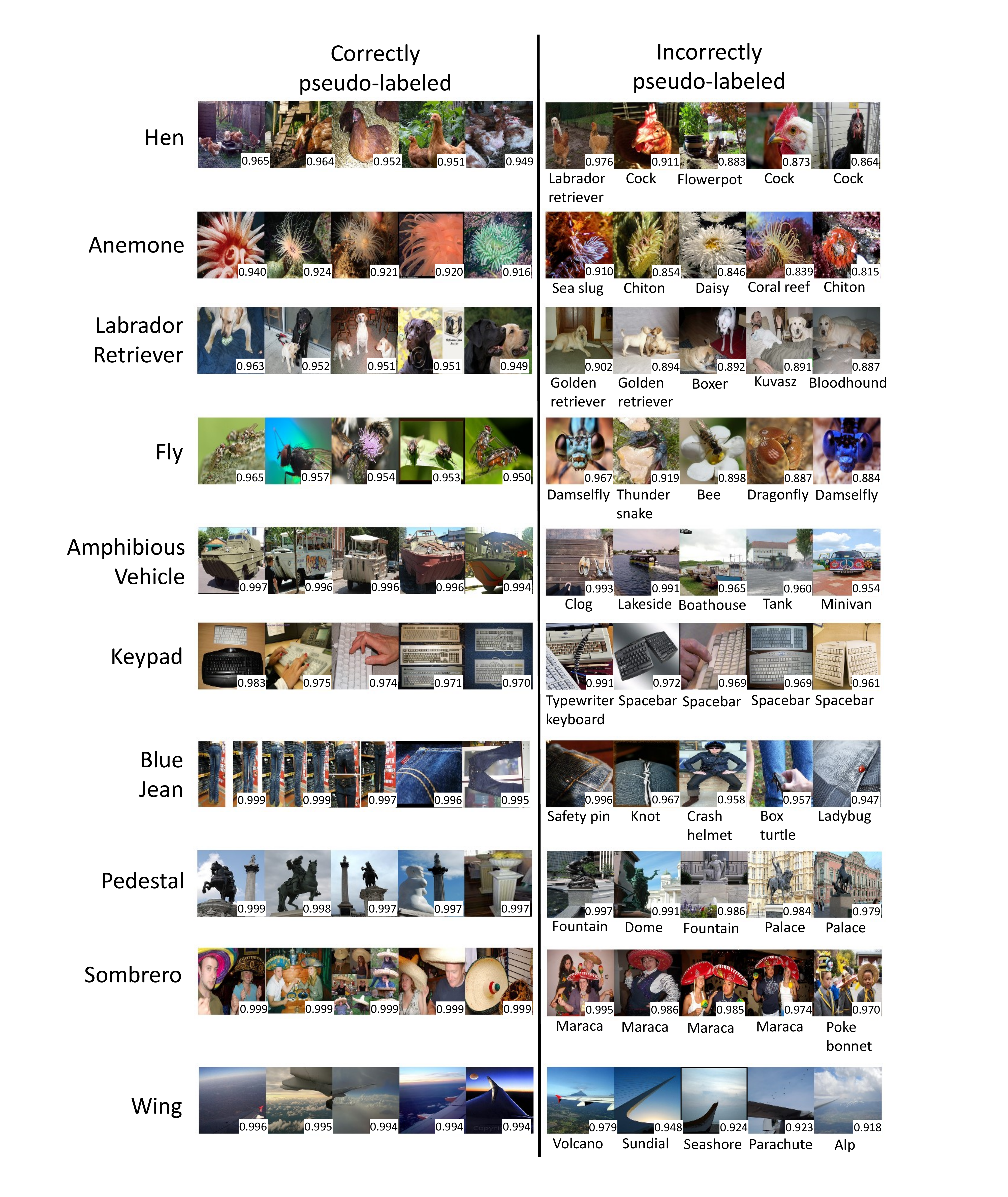}
\caption{ImageNet pseudo-labeled samples annotated by our method. We show correct and incorrect samples for 10 chosen categories with their corresponding scores after the last iteration.}
\label{fig:inSamples}
\end{figure*}

We show pseudo-labeled samples annotated by our method on CIFAR-10 (figure \ref{fig:cifarSamples}), and ImageNet (figure \ref{fig:inSamples}). We show correct and incorrect samples for 10 categories with their corresponding scores after the last iteration. We show the highest scores for each class in each column. Under the \textit{incorrectly pseudo-labeled} column we also show the ground truth category for the image annotated by the model.

\end{document}